\documentclass[10pt,twocolumn,letterpaper]{article}

\usepackage{cvpr}              % To produce the CAMERA-READY version
\usepackage{graphicx}
\usepackage{amsmath}
\usepackage{amssymb}
\usepackage{booktabs}

\usepackage[pagebackref,breaklinks,colorlinks]{hyperref}

\usepackage[capitalize]{cleveref}
\crefname{section}{Sec.}{Secs.}
\Crefname{section}{Section}{Sections}
\Crefname{table}{Table}{Tables}
\crefname{table}{Tab.}{Tabs.}

\def\cvprPaperID{*****} % *** Enter the CVPR Paper ID here
\def\confName{CVPR}
\def\confYear{2023}

\begin{document}

\title{Fact Sheet for CVPR 2023 Autonomous Driving Challenge\\
Track 2 Online HD Map Construnction}

% \author{
% Mingchao Jiang\\
% GAC R&D Center\\
% {\tt\small jiangshaoyu1993@gmail.com}
% \and
% Yin Cheng\\
% Beijing University of Posts and Telecommunications\\
% {\tt\small 3175280282@qq.com}
% \and
% Linghai Liu\\
% GAC R&D Center\\
% {\tt\small liulinghai9@gmail.com}
% }
\author{Mingchao Jiang$^{1}$, Yin Cheng$^{2}$, Linghai Liu$^{1}$\\
$^1$GAC R\&D Center, Guangzhou, China\\
$^2$ Beijing University of Posts and Telecommunications, Beijing, China\\
\small jiangshaoyu1993@gmail.com, 3175280282@qq.com, liulinghai9@gmail.com}
\maketitle

%%%%%%%%% BODY TEXT
\section{Team Name}
\label{sec:tn}
Team Name: MapSeg.

\setlength{\parindent}{1.2em} EvalAI User Name: @FlyEgle.
%------------------------------------------------------------------------
\section{Method}
\label{sec:method}
\noindent \textbf{Method Name:} Online High-precision Map Construction with Segmentation-guided Structured Modeling and Learning.

\begin{figure*}[t]
    \centering
    \includegraphics[width=1.0\textwidth]{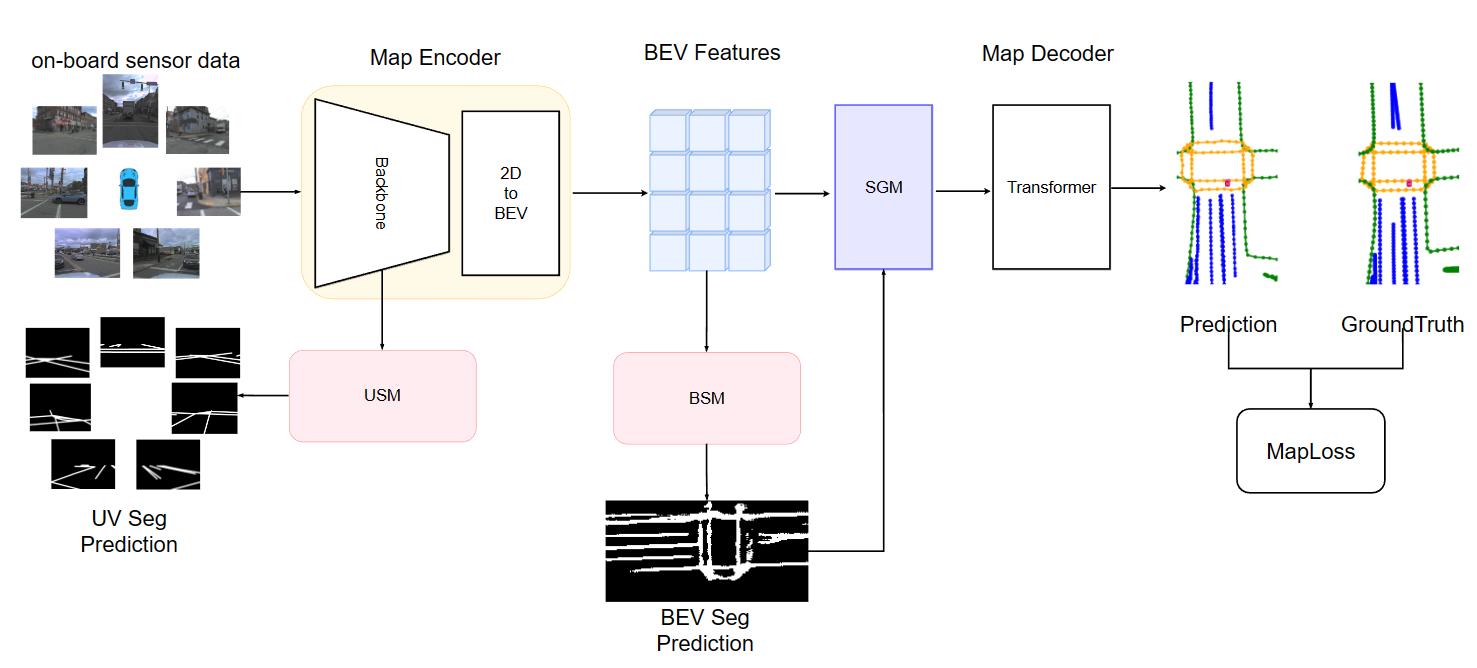}
    \caption{Overview of model architecture.The entire model is consists of by Map Encoder, USM(UV Segmentation Module), BSM(BEV Segmentation Module), SGM(Semantic Guidance Module), and Map Decoder.  }
%We give details of the structure and configurations in Section 3.}
    \label{fig: overview}
\end{figure*}

\noindent \textbf{Introduction: }
The development of online high-definition maps is significant since they provide real-time, accurate, and updatable geographic information for location-based applications, such as autonomous driving and intelligent transportation, thus improving the performance and reliability of these applications. Previous works, such as VectorMapNet\cite{liu2022vectormapnet} and MapTR\cite{liao2022maptr}, show that direct model generation of vectorized HD maps is a promising solution. However, these methods did not take into account the usage of global semantic information to improve map construction accuracy. To address this limitation, we propose a segmentation-guided structured model (MapSeg) for online HD map construction, as depicted in Figure \ref{fig: overview}. Specifically, we added a UV segmentation module (USM) and a BEV segmentation module (BSM) based on the MapTR structure, enabling the model to better capture the semantic information. What's more, to further improve the model's vectorization ability, we proposed a semantic guidance module (SGM). More details of USM, BSM and SGM modules are described as follows. The source code are available at \url{https://github.com/FlyEgle/CVPR_hdmap}. 

%-------------------------------------------------------------------------
\subsection{UV Segmentation Module}
In order to enhance the semantic capability of the model, we added a UV segmentation module on top of an FPN structure. The segmentation head from DeepLabV3\cite{chen2018encoder} is exploited directly as our USM and it is only effective during the training phase. This process is formulated as: 
\begin{equation}\label{eq1}
    \mathbf{O_{uv}} = \mathbf{USM}(\mathbf{FPN}(\mathbf{Backbone}(\mathbf{X_{img)}}))
\end{equation}
where $\mathbf{X_{img}}$ is surround image view. $\mathbf{O_{uv}}$ is uv segmentation feature.

%-------------------------------------------------------------------------
\subsection{BEV Segmentation Module}
To further improve the vector results generated during the BEV stage by the model, a BEV segmentation module is incorporated after the BEV features, which is also based on DeepLabV3. We call this module BSM. This process is formulated as:
\begin{equation}\label{eq2}
    \mathbf{O_{bev}} = \mathbf{BSM}(\mathbf{X_{bev}})
\end{equation}
where $\mathbf{X_{bev}}$ is the BEV feature after encoding, $\mathbf{O_{bev}}$ is the BEV segmentation feature.
%-------------------------------------------------------------------------
\subsection{Semantic Guidance Module}
To make the BEV features be more discriminative, we followed the structure of cross attention, so that the BEV segmentation module can better guide the extraction of the BEV feature. To ensure the completeness of the BEV information, we concatenate the guided BEV features with the original BEV features. This process is formulated as:
\begin{equation}\label{eq3}
    \mathbf{G_{bev}} = \mathbf{softmax}(\frac{f_{Q}(\mathbf{O_{bev}})f_{K}(\mathbf{X_{bev}})}{\sqrt{d_{k}}})f_{V}(\mathbf{X_{bev}})
\end{equation}
\begin{equation}\label{eq4}
    \mathbf{Y_{bev}} = \mathbf{Cat}[\mathbf{G_{bev}}, \mathbf{X_{bev}}]
\end{equation}
where $f_{Q}$,$f_{K}$,$f_{V}$ is the project function for $Q$,$K$,$V$. $\mathbf{G_{bev}}$ is the
guided BEV features and $\mathbf{Y_{bev}}$ is the concatenated BEV features.
%-------------------------------------------------------------------------
\subsection{Loss Function}
Given the UV segmentation gt $\mathbf{I_{uv}^{gt}}$, the output from the USM module is defined as $\mathbf{I_{uv}^{usm}}$. Given the BEV segmentation gt $\mathbf{I_{bev}^{gt}}$, the output from the BSM module is defined as $\mathbf{I_{bev}^{bsm}}$. We optimize this module with the following loss function:
\begin{equation}\label{eq5}
     \mathcal{L}_{usm} = \lambda_{1}\mathcal{L}_{Dice}(\mathbf{I_{uv}^{usm}}, \mathbf{I_{uv}^{gt}}) + \lambda_{2} \mathcal{L}_{CE}(\mathbf{I_{uv}^{usm}}, \mathbf{I_{uv}^{gt}})
\end{equation}
\begin{equation}\label{eq6}
     \mathcal{L}_{bsm} = \lambda_{1}\mathcal{L}_{Dice}(\mathbf{I_{uv}^{bsm}}, \mathbf{I_{uv}^{gt}}) + \lambda_{2} \mathcal{L}_{CE}(\mathbf{I_{uv}^{bsm}}, \mathbf{I_{uv}^{gt}})
\end{equation}
\begin{equation}\label{eq7}
    \mathcal{L}_{seg} = \mathcal{L}_{usm} + \mathcal{L}_{bsm}
\end{equation}
where $\mathcal{L}_{CE}$ is the Cross-Entropy loss, and $\mathcal{L}_{Dice}$ is the Dice loss:
\begin{equation}\label{eq8}
    \mathcal{L}_{Dice} = 1 - 2\cdot\frac{pred \bigcap true}{pred \bigcup true}
\end{equation}
The total loss of the model can be expressed as:
\begin{equation}\label{eq9}
    \mathcal{L}_{total} = \mathcal{L}_{maptr} + \mathcal{L}_{seg}
\end{equation}
where $\mathcal{L}_{maptr}$ is the MapTR loss function.

\subsection{Training}
\noindent \textbf{Model Setting: }We generally follow the MapTR-Tiny structure, where the encoder-layer uses one layer and the decoder-layer uses six layers. For DeepLabV3, we follow the structure implemented by the mmsegmentation framework. Considering to reduce the learning difficulty, our semantic segmentation module only predicts two categories, foreground and background.

\noindent \textbf{Data Augmentation: }We merge the training data and validation data for training. In the training stage, since the surround image size is not the same, we resized the input images to $1600 \times 2048$ and adjusted the camera intrinsic as well. We performed random horizontal flipping, random rotation, and color distortion to augment the training data.

\noindent \textbf{Training Strategy: }Our model was optimized with the AdamW\cite{Alpher06} method with $\beta_{1}=0.9, \beta_{2}=0.999$ and a batch size of 2. Our model was implemented using PyTorch\cite{Alpher07} and trained with 4 NVidia A100 GPUs. The learning rate was initially set to $3 \times 10^{-4}$ and scheduled with the cosine annealing strategy. ResNet101\cite{he2016deep} was adopted as our backbone, and trained for 100 epochs. To reduce the model inference latency, we use the ResNet50\cite{he2016deep} replace the ResNet101 as the backbone. To ensure the accuracy of the model, we used the ResNet101 best model as the ResNet50 pretrain. Finally, we training 30 epochs with ResNet50 as our results.For $\mathcal{L}_{seg}$ the $\lambda_{1}$ set to 15, $\lambda_{2}$ set to 0.5. 
\subsection{Testing}
We did not perform any model ensembling or test time augmentation (TTA) during the testing phase. In the end, the ResNet50 as backbone was used for testing.

%%%%%%%%% Experiments
\section{Ablation Study}
In this section, we presented the performance changes of the proposed different modules on the validation dataset. The baseline is used the ResNet50.

In the table \ref{tab:module}, we compared the ablation experiments of different combinations of USM, BSM, and SGM,All module backbone is used the ResNet50.The results showed that all three modules have a significant gain on the original model structure. 

In the table \ref{tab:parameters}, we found that the larger resolutions and the larger models can further improve the performance of the model.

% table with mapseg ablation
\begin{table}[htbp]
 \centering
 \resizebox{8.5cm}{1.2cm}{
 \begin{tabular}{lcccccl}\toprule
    % \\\cmidrule(lr){2-4}\cmidrule(lr){5-7}
        Module     & size  & ped crossing & divider    &boundary   &map  \\\midrule
    baseline    & $800\times 1024$ & 0.5190 & 0.6162  & 0.6047 & 0.5800 \\
    USM & $800\times 1024$ & 0.5419 & 0.6185 & 0.6219 & 0.5941 \\
    BSM & $800\times 1024$ & 0.5381 & 0.6249 & 0.617 & 0.5933 \\
    USM + BSM  & $800\times 1024$ & \textbf{0.5442}  &0.6405 & 0.6317 & 0.6054  \\
    USM + BSM + SGM &$800\times 1024$ &0.5397 &\textbf{0.6512} &\textbf{0.6419} &\textbf{0.6109}
    \\\bottomrule
 \end{tabular}}
 \caption{\centering The module ablation experiments of USM, BSM, SGM.To boldface indicates the best results.}
 \label{tab:module}
\end{table}

% table with parameters ablation
\begin{table}[htbp]
 \centering
 \resizebox{8.5cm}{1cm}{
 \begin{tabular}{lcccccl}\toprule
    % \\\cmidrule(lr){2-4}\cmidrule(lr){5-7}
        backbone  &size    & ped crossing & divider    &boundary   &map  \\\midrule
    ResNet50    & $800\times 1024$ & 0.5397 & 0.6512  & 0.6419 & 0.6109 \\
    ResNet50 & $1600\times 2048$ & 0.5532 & 0.6685 & 0.6538 & 0.6294 \\
    ResNet101 & $1600\times 2048$ & \textbf{0.5714} & \textbf{0.6824} & \textbf{0.6647} & \textbf{0.6452}
    \\\bottomrule
 \end{tabular}}

\caption{\centering{The module ablation experiments for resolution \& model parameters.}}
    
    \label{tab:parameters}
\end{table}

%%%%%%%%% Members
\section{Members}
    Mingchao Jiang (jiangshaoyu1993@gmail.com) 

    \setlength{\parindent}{1.2em}Yin Cheng (3175280282@qq.com)
    
    \setlength{\parindent}{1.2em}Linghai Liu (liulinghai9@gmail.com)
    
    \setlength{\parindent}{1.2em}\textit{The first member will be referred to as the captain of the team.}

%%%%%%%%% Affliation
\section{Affliation}
GAC R\&D Center, Beijing University of Posts and Telecommunications.

%%%%%%%%% REFERENCES
{\small
\bibliographystyle{ieee_fullname}
\bibliography{main}
}

\end{document}